\documentclass[10pt,twocolumn,letterpaper]{article}

\usepackage{iccv}
\usepackage{times}
\usepackage{epsfig}
\usepackage{graphicx}
\usepackage{amsmath}
\usepackage{amssymb}
\usepackage{tabularx}
\usepackage{float}

\usepackage{xcolor}
\usepackage{caption}

\usepackage[pagebackref=true,breaklinks=true,letterpaper=true,colorlinks,bookmarks=false]{hyperref}

\iccvfinalcopy 

\def\iccvPaperID{4864} 

\ificcvfinal\fi

\title{Towards Accurate Generative Models of Video: A New Metric \& Challenges}

\author{
\makebox[.25\linewidth]{Thomas Unterthiner\footnotemark}\\
Johannes Kepler University\\
{\tt\small unterthiner@ml.jku.at}
\and
\makebox[.23\linewidth]{Sjoerd van Steenkiste$^{*}$}\\
IDSIA, SUPSI, USI\\
{\tt\small sjoerd@idsia.ch}
\and
\makebox[.25\linewidth]{Karol Kurach}\\
Google Brain\\
{\tt\small kkurach@google.com}
\and
\makebox[.25\linewidth]{Rapha\"{e}l Marinier}\\
Google Brain\\
{\tt\small raphaelm@google.com}
\and
\makebox[.25\linewidth]{Marcin Michalski}\\
Google Brain\\
{\tt\small michalski@google.com}
\and
\makebox[.25\linewidth]{Sylvain Gelly}\\
Google Brain\\
{\tt\small sylvaingelly@google.com}
}

\begin{document}

\twocolumn[{%
\renewcommand\twocolumn[1][]{#1}%
\maketitle
\begin{center}
    \centering
    \includegraphics[width=\textwidth]{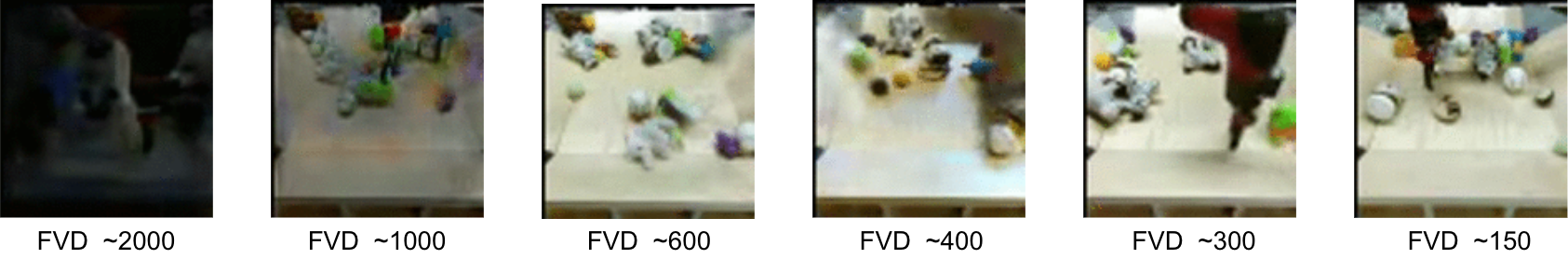}
    \captionof{figure}{Generated videos by various models ranked according to FVD (lower is better) on the BAIR dataset~\cite{bib:ebert2017:bair}.}
    \label{fig:fvdbair}
\end{center}%
}]



\begin{abstract} \let\thefootnote\relax\footnote{$^*$Both authors contributed equally to this work while interning at Google Brain.}\addtocounter{footnote}{-1}\let\thefootnote\svthefootnote 
    Recent advances in deep generative models have lead to remarkable progress in synthesizing high quality images.
    Following their successful application in image processing and representation learning, an important next step is to consider videos.
    Learning generative models of video is a much harder task, requiring a model to capture the temporal dynamics of a scene, in addition to the visual presentation of objects.
    While recent attempts at formulating generative models of video have had some success, current progress is hampered by (1) the lack of qualitative metrics that consider visual quality, temporal coherence, and diversity of samples, and (2) the wide gap between purely synthetic video data sets and challenging real-world data sets in terms of complexity. 
    To this extent we propose Fr\'{e}chet Video Distance (FVD), a new metric for generative models of video, and StarCraft 2 Videos (SCV), a benchmark of game play from custom starcraft 2 scenarios that challenge the current capabilities of generative models of video.
    We contribute a large-scale human study, which confirms that FVD correlates well with qualitative human judgment of generated videos, and provide initial benchmark results on SCV.
\end{abstract}


\section{Introduction}
Recent advances in deep generative models have lead to remarkable success in synthesizing high-quality images~\cite{bib:karras2018:progressivegrowing, bib:brock2018:biggan}.
Their versatility offers a unified approach to a variety of image processing and computer vision tasks.
For example, Generative Adversarial Networks (GANs;~\cite{bib:goodfellow2014:gans}) may be used to perform image super-resolution~\cite{bib:sajjadi2017:enhancenet}, image-to-image translation~\cite{bib:isola2017:img2img}, and semantic segmentation~\cite{bib:luc2016:semantic}. 
Similarly, other generative approaches are able to extract useful representations of an image via inference, and have shown promising results in semi-supervised learning~\cite{bib:radosavovic2018:omnisupervised}, and few-shot learning~\cite{bib:lake2017}.

An important next challenge is to learn generative models of video, which requires that models capture the \emph{temporal dynamics} of a visual scene, e.g.\  how objects interact, in addition to their visual presentation.
They are expected to facilitate a wide range of applications, including missing-frame prediction~\cite{bib:jiang2018:superslomo}, improved instance segmentation~\cite{bib:haller2017}, or complex (relational) reasoning tasks by conducting inference~\cite{bib:lerer2016}.
Historically, many different approaches to video generation have been explored.
Traditional approaches train a recurrent neural network (e.g.\ an LSTM~\cite{bib:hochreiter1997:lstm}) to perform next-frame prediction. 
These models capture temporal dependencies through their recurrent connections, which can be iteratively applied to obtain future frames from a particular state~\cite{bib:ranzato2014, bib:srivastava2015, bib:mathieu2016, bib:Finn2016:cdna, bib:lotter2017:prednet}.
However, for a given set of context frames, the resulting output sequence is deterministic, which fails to account for the many possible futures that a sequence of context frames may have (e.g. due to external, unobserved factors).
More recent approaches either additionally condition on a set of external random variables (latent factors)~\cite{bib:babaeizadeh2017:sv2p, bib:denton2018:svg, bib:lee2018:savp}, generate a sequence of observation entirely from noise, e.g. using GANs~\cite{bib:vondrick2016:vgan,bib:saito2017:tgan,bib:tulyakov2017:mocogan} or factorize the joint distribution over the whole video, such that each pixel in the sequence is conditioned on all previously generated pixels~\cite{bib:kalchbrenner2017:vpn, bib:byeon2018:contextvp}.

While great progress has been made in recent years, video generation models are still in their infancy, and generally unable to synthesize more than a few seconds of video~\cite{bib:babaeizadeh2017:sv2p, bib:villegas2018:hierarchical}.
Indeed, by looking at generated samples it becomes clear that learning a good dynamics model remains a major challenge. 
However, in order to \emph{qualitatively} measure progress in accurately synthesizing videos we need corresponding \mbox{\emph{metrics}} that consider visual quality, temporal coherence, and diversity of generated samples.
Likewise, in order to isolate (independent) factors that contribute to different failure modes, we require corresponding \emph{data sets} that test for specific capabilities. 
In this work we present the following contributions to address these challenges:

\begin{itemize}
\item We introduce \emph{Fr\'{e}chet Video Distance (FVD)}, a new metric for generative models of video\footnote{Code to compute FVD is available at \url{https://git.io/fpuEH}.}$^{\ref{note1}}$.
FVD builds on the principles underlying Fr\'{e}chet Inception Distance (FID;~\cite{bib:heusel2017:fid}), which has been successfully applied to images, and was later adapted to other fields~\cite{bib:Preuer2018:fcd}.
We introduce a different feature representation that captures the temporal coherence of a video, in addition to the quality of each frame.
Unlike popular metrics such as the Peak Signal to Noise Ratio (PSNR) or the Structural Similarity (SSIM;~\cite{bib:wang2004:ssim}) index, FVD considers a distribution over entire videos, thereby avoiding the drawbacks of frame-level metrics~\cite{bib:huynh-thu2012:psnr}.
\item We investigate several variants of FVD by considering multiple distance functions and embeddings. By adding noise to real videos we show that FVD is sensitive to both temporal, and frame-level perturbations.
\item We perform a large-scale human study that confirms that FVD coincides well with qualitative human judgment of generated videos.

\item We introduce \emph{StarCraft 2 Videos (SCV)}, a suite of challenging data sets that require relational reasoning and long-term memory\footnote{SCV is available at \url{https://goo.gl/HJuQPf}, or through \url{https://www.tensorflow.org/datasets}.}.
Using the open-source StarCraft 2 Learning Environment (SC2LE;~\cite{bib:vinyals2017:sc2le}), we contribute four data sets consisting of game play from handcrafted StarCraft 2 scenarios.
Each data set is scalable along many axes, including the resolution and complexity of the scenes.
\item We provide a very comprehensive comparison of current state of the art models on BAIR, KTH and SCV datasets in terms of FVD. By varying hyper-parameters and resolution, our study examines $3000$ models. In total, our experiments amount to over $100$ GPU-years worth of computation.
\end{itemize}
\addtocounter{footnote}{1}\footnotetext{\label{note1}A similar adaptation of FID was used by Wang et al.~\cite{bib:wang2018:vid2vid} to evaluate their \emph{vid2vid} model. Here we introduce FVD as a general metric for videos and an extensive empirical study of its capabilities is the focus of our work.}

\section{Fr\'{e}chet Video Distance}\label{sec:fvd}
An accurate generative model of videos captures the data distribution from which the observed data was generated.
Hence, the distance between the real world data distribution $P_R$ and the distribution defined by the generative model $P_G$ is an obvious evaluation metric.
Unfortunately no analytic expression of either distribution is available, which rules out straightforward application of many common distance functions.
Consider for example the popular Fr\'{e}chet Distance (or 2-Wasserstein distance) between $P_R$ and $P_G$ defined by:
\begin{equation} \label{eq:w2}
    d(P_R, P_G) = min_{X, Y} \mathrm{E} |X - Y|^2
\end{equation}
where the minimization is over all random variables X and Y with distributions $P_R$ and $P_G$ respectively.
This expression is difficult to solve for the general case, although it has a closed form solution when $P_R$ and $P_G$  
are multivariate Gaussians~\cite{bib:dowson:frechet}. In that case the right-hand side in Eq.~\ref{eq:w2} reduces to:
\begin{align} \label{eq:fid}
    |\mu_R - \mu_G|^2 + \mathrm{Tr}\left(\Sigma_R + \Sigma_G - 2 (\Sigma_R \Sigma_G)^{\frac{1}{2}} \right)
\end{align}
where $\mu_R$ and $\mu_G$ are the means and $\Sigma_R$ and $\Sigma_G$ are the co-variance matrices of $P_R$ and $P_G$ respectively.

Thus, evaluating Eq.~\ref{eq:w2} becomes feasible if we assume a particular form of the distributions under consideration.
A multivariate Gaussian is seldom an accurate representation of the underlying data distribution, but when using a suitable feature space, it is a reasonable approximation.
For distributions over real world \emph{images}, Heusel et al.~\cite{bib:heusel2017:fid} used a learned feature embedding to calculate the distance between $P_R$ and $P_G$ as follows:
First, an Inception network~\cite{bib:szegedy2016:inceptionv3} is trained on ImageNet~\cite{bib:deng2009:imagenet} to classify images.
Next, samples from $P_R$ and $P_G$ are fed through the pre-trained network and their feature representation (activations) in one of the hidden layers is recorded.
Finally, the Fr\'{e}chet Inception Distance (FID; \cite{bib:heusel2017:fid}) is computed according to Eq.~\ref{eq:fid} using the means and covariances obtained by fitting a multivariate Gaussian distribution to the recorded responses of the real, and generated samples.

The feature representation learned by the pre-trained neural network greatly affects the quality of the metric.
By training on ImageNet, a model mainly focuses on the objects in images, potentially suppressing other information content.
Likewise, different layers of the network encode features at different abstraction levels.
A suitable feature representation for videos needs to consider the temporal coherence of the visual content across a sequence of frames, in addition to its visual presentation at any given point in time.
In this work we investigate several variations of a pre-trained Inflated 3D Convnet (I3D;~\cite{bib:carreira2017:id3kinetics400}), and name the resulting metric the \emph{Fr\'{e}chet Video Distance (FVD)}.
The I3D network generalizes the Inception architecture to sequential data, and is trained to perform action-recognition on the Kinetics data set consisting of human-centered YouTube videos~\cite{bib:kay2017:kinetics}. 
Action-recognition can be viewed as a temporal extension of image classification, requiring visual context and temporal evolution to be considered simultaneously.
I3D has been shown to excel at this task: it achieved state-of-the-art results on the UCF101\cite{bib:soomro2012:ucf101} and HMDB51~\cite{bib:kuehne2013:hmdb51} data sets, and placed first at the CVPR 2017 Charades challenge~\cite{bib:kay2017:kinetics}.
We consider I3D networks that have been trained on the RGB frames of the Kinetics-400 data set, and on the Kinetics-600 data set\footnote{pre-trained weights that are available at \url{https://tfhub.dev/}.}.
In order to obtain suitable feature representations we evaluate the logits in the final layer, as well as the output of the last pooling layer analogous to prior work~\cite{bib:heusel2017:fid}.
\begin{figure*}[h!]
\begin{center}
\includegraphics[width=1.0\linewidth]{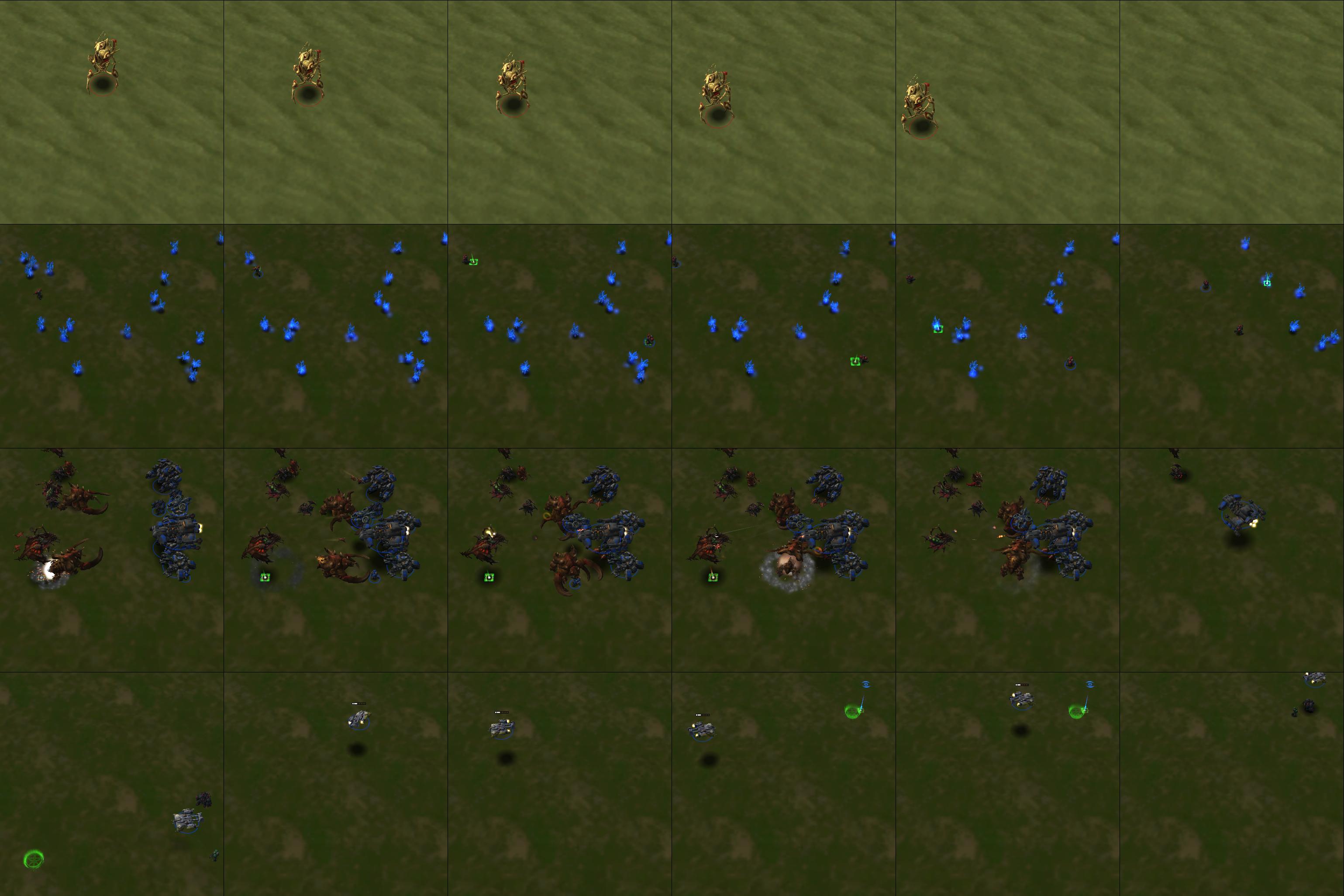}
\end{center}
\caption{Video frames of each SCV scenario sampled at regular intervals.
         From top to bottom: \emph{Move Unit to Border (MUtB)}, \emph{Collect Mineral Shards (CMS)}, \emph{Brawl}, and \emph{Road Trip with Medivac (RTwM)}.}
\label{fig:scv}
\end{figure*}

A potential downside of using Eq.~\ref{eq:fid} is the potentially large error in estimating Gaussian distributions over the learned feature space.
As an alternative, Binkowski et al. \cite{bib:binkowski2018:kid} proposed to use the Maximum Mean Discrepancy (MMD~\cite{bib:gretton2012:mmd}) in the case of images, and we will explore this variation in the context of videos as well.
MMD is a kernel-based approach, which provides a means to calculate the distance between two empirical distributions without assuming a particular form.
Concretely, if ${x_1 \ldots x_m} $ and ${y_1 \ldots y_n}$ are samples drawn from two random variables $X$ and $Y$ with distributions $P_R$ and $P_G$, then an unbiased estimator of the squared MMD distance between these two random variables is given as:
\begin{align*} \label{eq:mmd}\tiny
\sum_{i\neq j}^{m} \frac{k(x_i, x_j)}{m(m-1)} - 2\sum_i^m \sum_j^n \frac{k(x_i, y_j)}{m n} + \sum_{j\neq j}^{n} \frac{k(y_i, y_j)}{n(n-1)} 
\end{align*}
where $k(\cdot, \cdot)$ is a kernel that measures the similarity between two input vectors. 
Binkowski et al.~\cite{bib:binkowski2018:kid} proposed to use a polynomial kernel $k(a, b):= \left( a^Tb + 1\right)^3$, which we will apply to the learned features of the I3D network to obtain the Kernel Video Distance (KVD).

In our experiments in Section~\ref{sec:experiments}, we will compare FVD (KVD) to the two main metrics that are currently used in the relevant literature: PSNR, and SSIM.
The Peak Signal to Noise Ratio (PSNR) relates the maximum attainable pixel value of the pixels in an image to its Mean Squared Error (MSE) with respect to a ground-truth image.
In the case of videos, the PSNR is computed for each frame with respect to a reference frame.
Several studies have pointed out that PSNR does not correlate well with subjective video quality, in particular when simultaneously evaluating multiple videos with different content~\cite{bib:huynh-thu2012:psnr}.
The Structural Similarity (SSIM;~\cite{bib:wang2004:ssim}) index measures the quality of an image as the perceived change in structural information.
Similar to the PSNR, the SSIM metric requires access to ground-truth frames, and considers each frame individually.
It suffers from the same drawbacks in comparing distributions over videos (with potentially different content) according to human judgment~\cite{bib:ponomarenko2015, bib:Zhang2018}.

An advantage of SSIM and PSNR is that they can be used to measure the degree to which a specific generated sequence deviates from a ground-truth sequence, i.e. given a particular context.
FVD necessarily considers only \emph{distributions} of videos, although we will show in Section~\ref{sec:experiments} (Switching noise) that in combining the context frames and the generated frames it can account for generated frames that do not necessarily progress from a sequence of context frames.
On the other hand, FVD applies naturally to the unconditional case in which no ground-truth sequences are available (e.g. GANs), and estimates the variety among the generated sequences in addition to their visual and temporal coherence. It is impossible to apply PSNR or SSIM in such scenarios.

\section{Starcraft 2 Videos}
In order to advance the capabilities of generative models for video, a number of challenges must be overcome.
For example, generative models must learn that similar actions might be performed by agents differing in their visual appearance (e.g. KTH \cite{bib:schuldt2004:kth}), and learn to account for the highly non-linear temporal dynamics in observing a robot arm that pushes objects around on a table~\cite{bib:Finn2016:cdna}. 
Here we propose to consider these challenges in a simpler setting (the StarCraft 2 Learning Environment~\cite{bib:vinyals2017:sc2le}) to serve as an intermediate step towards real world video data sets.

We introduce StarCraft 2 Videos (SCV), a suite of benchmark data sets for video generation.
SCV offers tasks that are meant to serve as `unit tests' while developing new models, and more challenging tasks that are meant to bring current models to their limits.
The latter isolate specific challenges in real-world video generation, such as long-term memory or relational reasoning, and emphasize typical failure modes and avenues for future research.
Each SCV data set is generated according to a predefined scenario (including many stochastic choices), which can be easily adapted to alter its complexity.
Videos are obtained by rendering scenes from the video game StarCraft 2 (SC2), using the open-source StarCraft 2 Learning Environment (SC2LE;~ \cite{bib:vinyals2017:sc2le}), and can be rendered at different resolutions.
SCV includes four different scenarios, for which we can generate a virtually unlimited number of videos.
Example frames from each scenario can be seen in Figure~\ref{fig:scv}.

\paragraph{Move Unit to Border (MUtB)} 
A randomly chosen game unit moves from the middle of the map to its border. This introductory test scenario offers similar complexity to ``Moving MNIST''~\cite{bib:srivastava2015} or ``Stochastic Shapes''~\cite{bib:babaeizadeh2017:sv2p} in terms of temporal dynamics, yet is visually more complex.
In addition to modeling several stochastic dimensions (i.e. the type of game unit, its color, and destination along the border) a model must also learn about the animation sequence associated with each moving unit.

\paragraph{Collect Mineral Shards (CMS)} 
Two units move across the screen to collect randomly placed mineral shards in a greedy fashion\footnote{This map, and the corresponding agent, were originally developed as part of the original SC2LE framework~\cite{bib:vinyals2017:sc2le}.}.
In terms of dynamics it requires a video model to learn the relation between a unit reaching a shard and the shard disappearing, and about the shortest-path pattern in which units walk from location to location.

\paragraph{Brawl} 
Two large armies (army compositions and starting location vary across videos) face each other and attempt to destroy the other. 
The ``Terran'' (left) army is ordered to attack, repeatedly targeting the nearest enemy unit, such that the winner is solely determined by the army composition, and initial location of units. 
This scenario offers highly complex temporal interactions, requiring an accurate generative model to learn about properties of units (i.e. type, health points, attack range, attack damage) and model their interactions, various (attack) animations, and the many small objects (e.g. rockets) flying across the scene.

\begin{figure*}[ht!]
\begin{center}
\includegraphics[width=1.0\linewidth]{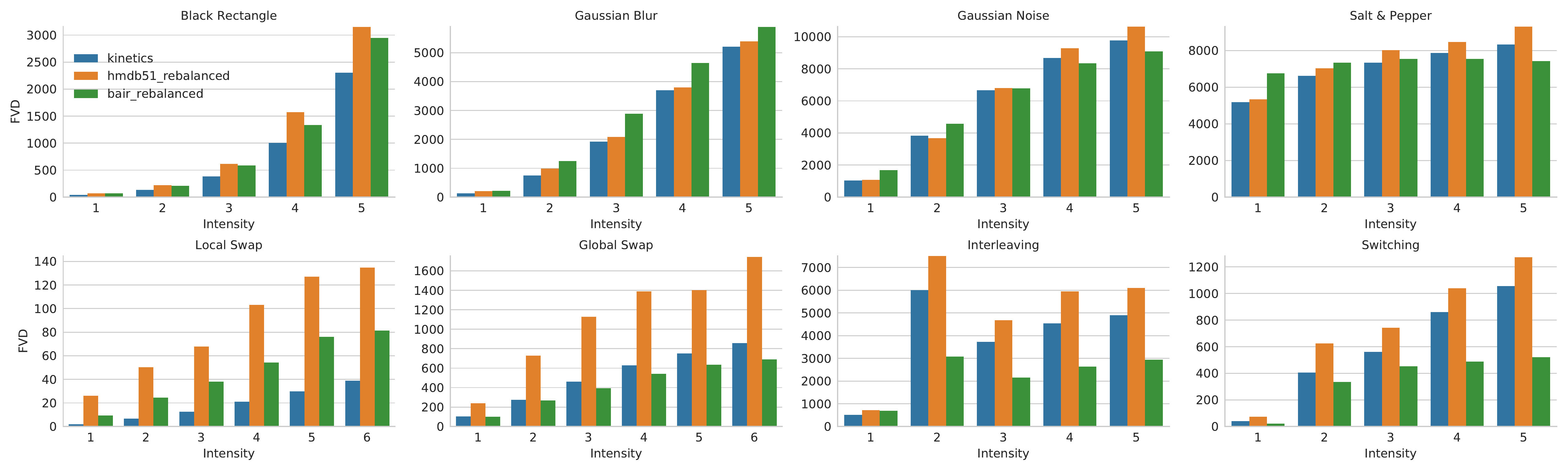}
\end{center}
\caption{Behaviour of FVD when adding various types of noise to different data sets, using the logits activations of the I3D model trained on Kinetics-400 as embedding.}
\label{fig:basicnoise}
\end{figure*}

\paragraph{Road Trip with Medivac (RTwM)} 
A flying transportation unit (``medivac'') is tasked to pick up a small group of units (one by one) that spawn at nearby locations. 
Next, it is tasked to visit a number of beacons, before eventually unloading the exact same units at the final beacon.
Modeling long-term dependencies is critical to succeed at this dataset requiring an understanding of what invidual units are, their types, and their interaction with the medivac.

The gameplay in each scenario is conceptually simple, which makes it straightforward to determine whether a generative model has managed to learn it. Eg. any valid RTwM video must consist of a pick-up, transportation, and drop-off phase, and the same number and type of units that were picked up must be dropped off again.  In contrast, in real world data sets (eg. BAIR, Kinetics) this is often hard to determine success by visual inspection alone. 
The ability to control the content of each video comes at a cost, and the visual fidelity of SCV can not rival the real world.
However, we note that the Starcraft 2 game engine offers many complex non-rigid motions, especially at higher resolutions, which we believe to be already sufficiently challenging for current models.
Meanwhile, SCV aims to provide a test bed for modelling interactions among multiple entities and consistency over long time horizons.
These are the skills required to solve SCV, and precisely what current video models struggle at~\cite{bib:babaeizadeh2017:sv2p}.

Each SCV scenario has several hyperparameters, which can be altered to increase or decrease its complexity.
We provide default parameters (details in the Appendix) for each scenario and supply corresponding data sets in two different resolutions ($64 \times 64$ and $128 \times 128$).
For these values we find (Section~\ref{sec:benchmark}) that state-of-the-art models only achieve moderate results on MUtB, and are unable to accurately capture the complex temporal dynamics (and corresponding consistency) in all other scenarios.
Nevertheless, we also release code to generate data sets with custom hyperparameters for each SCV scenario, to facilitate future generations of video models.

\section{Experiments}\label{sec:experiments}
\subsection{Noise Study}\label{sec:fvd-noisestudy}

We test whether FVD is sensitive to a number of basic distortions by adding various types of noise to real videos.
We consider \emph{static} noise added to individual frames, and \emph{temporal} noise, which distorts the entire sequence of frames.
While common image-based metrics are capable of detecting static noise, they were not designed to detect temporal noise.

To test whether FVD can detect static noise we added one of the following distortions to each frame in a sequence of video frames: (1) a \emph{black rectangle} drawn at a random location in the frame, (2) \emph{Gaussian blur}, which applies a Gaussian smoothing kernel to the frame, (3) \emph{Gaussian noise}, which interpolates between the observed frame and standard Gaussian noise, and (4) \emph{Salt \& Pepper noise}, which sets each pixel in the frame to either black or white with a fixed probability.
Temporal noise was injected by (1) \emph{locally swapping} a number of randomly chosen frames with its neighbor in the sequence (2) \emph{globally swapping} a number of randomly chosen pairs of frames selected across the whole sequence, (3) \emph{interleaving} the sequence of frames corresponding to multiple different videos to obtain new videos, and by (4) \emph{switching} from one video to another video after a number of frames to obtain new videos.
We applied these distortions at up to six different intensities that are unique to each type, e.g. related to the size of the black rectangle, the number of swaps to perform, or the number of videos to interleave.
Details are available in the Appendix.

We computed the FVD and KVD between videos from the BAIR~\cite{bib:ebert2017:bair}, Kinetics-400~\cite{bib:carreira2017:id3kinetics400} and HMDB51\cite{bib:kuehne2013:hmdb51} data sets and their noisy counterparts.
As potential embeddings, we considered the top-most pooling layer, and the logits layer of the I3D model pre-trained on the Kinetics-400 data set, as well as the same layers in a variant of the I3D model pre-trained on the extended Kinetics-600 data set.
As a baseline, we compared to a naive extension of FID for videos in which the Inception network (pre-trained on ImageNet~\cite{bib:deng2009:imagenet}) is evaluated for each frame individually, and the resulting embeddings (or their pair-wise differences) are averaged to obtain a single embedding for each video.
This ``FID'' score is then computed according to Eq.~\ref{eq:fid}.

We observed that all variants were able to detect the various injected distortions to some degree, with the pre-trained Inception network generally being inferior at detecting temporal distortions as was expected.
In Figure 1 of the Appendix
it can be seen that the logits layer of the I3D model pre-trained on Kinetics-400 is among the best performing configurations in terms of rank correlation with the sequence of noise intensities.
Hence, in the remainder of this work we will continue to use this configuration when computing FVD\footnote{A preliminary study in which a human ranking of generated videos was compared to the ranking obtained by each metric further corroborated this choice.}. 
An overview of its scores on the noise experiments can be seen in Figure~\ref{fig:basicnoise}.

\subsection{Effect of Sample Size on FVD}\label{sec:fvdsamplesize}
\begin{figure}[ht]
\begin{center}
\includegraphics[width=1.0\linewidth]{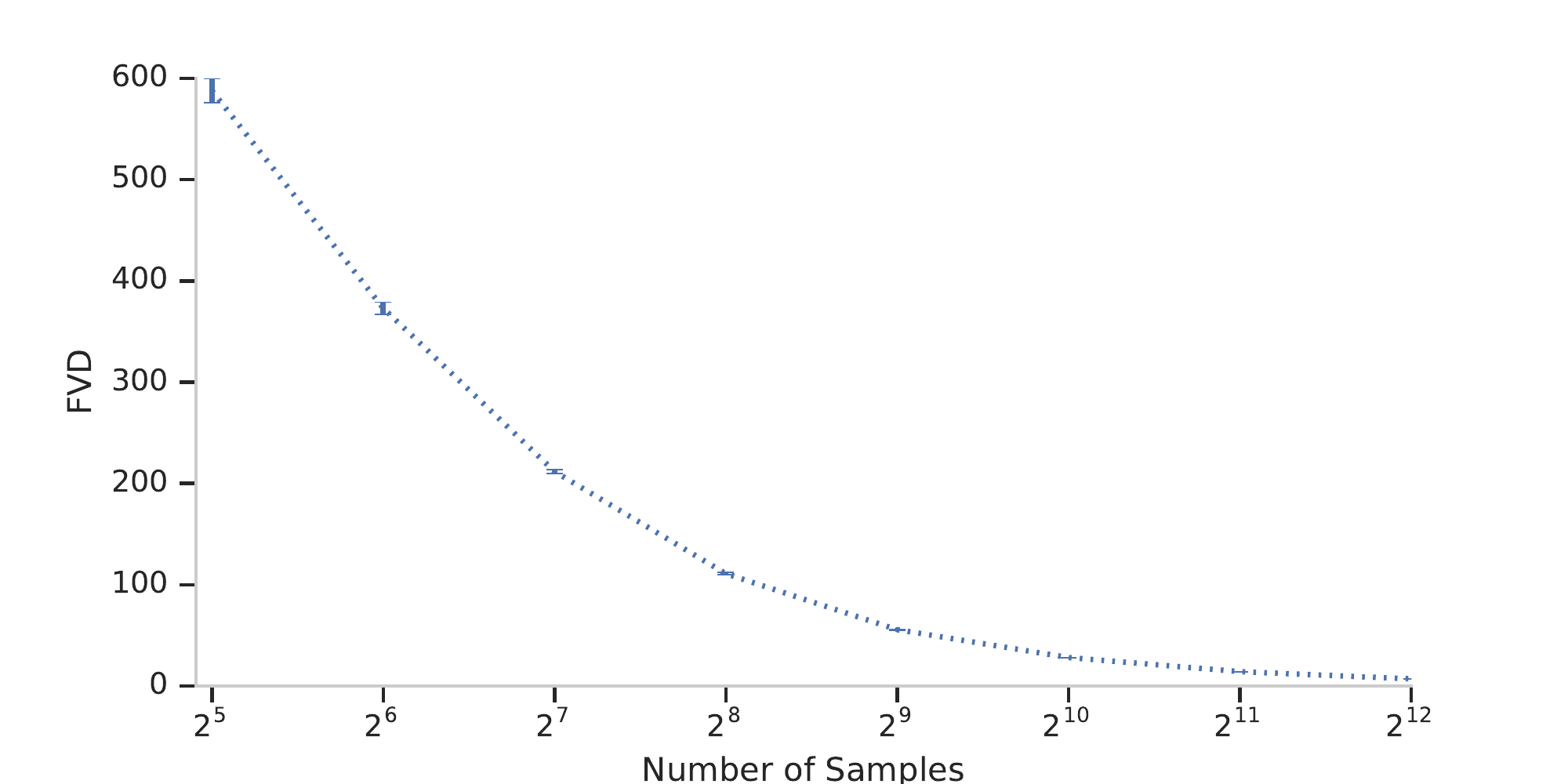}
\end{center}
\caption{FVD between two non-overlapping subsets of videos that are randomly drawn from the BAIR video pushing data set. Error bars are standard errors over 50 different tries.}
\label{fig:fvdbias}
\end{figure}

We consider the accuracy with which FVD approximates the true underlying distance between a distribution of generated videos and a target distribution.
To calculate the FVD according to Eq.~\ref{eq:fid} we need to estimate $\mu_R , \mu_G$ and $\Sigma_R, \Sigma_G$ from the available samples.
The larger the sample size, the better these estimates will be, and the better FVD will reflect the true underlying distance between the distributions.
For an accurate generative model these distributions will typically be fairly close, and only the noise from the estimation process is expected to affect our results.
This effect has been well-studied for FID~\cite{bib:lucic2018:comparegan, bib:binkowski2018:kid}, and is depicted for FVD in Figure~\ref{fig:fvdbias}.
It can be seen that even when the underlying distributions are identical, FVD will typically be larger than zero because our estimates of the parameters $\mu_R , \mu_G, \Sigma_R$ and $ \Sigma_G$ are noisy.
It can also be seen that for a fixed number of samples the standard errors (measured over 50 tries) are small, and an accurate comparison can be made.
Hence, it is critical that in comparing FVD values across models, one uses the same sample size\footnote{This has lead to confusion regarding FID in the past, when researchers used different sample sizes in their comparisons~\cite{bib:lucic2018:comparegan, bib:unterthiner2018:coulombgan, bib:heusel2017:fid}.}.

\subsection{Human Evaluation}\label{sec:fvd-humanstudy}

\begin{table*}[h]
\begin{center}
\begin{tabular}{l|c|c|c|c||c|c|c|c} \hline
\textbf{Metric} & \textbf{eq. FVD} & \textbf{eq. SSIM} & \textbf{eq. PSNR} & \textbf{eq. KVD} & \textbf{spr. FVD} & \textbf{spr. SSIM} & \textbf{spr. PSNR} & \textbf{spr. KVD} \\ \hline

FVD            & N/A	      & \textbf{74.9} \% & \textbf{81.0} \% & \textbf{63.0} \% & \textbf{71.9} \% & 58.4 \% & 63.5 \% & \textbf{63.1} \%\\
SSIM           & 51.5 \% & N/A       & 44.6 \% & 43.6 \% & 61.8 \% & 51.2 \% & 45.9 \% & 50.2 \% \\
PSNR           & \textbf{56.3} \% & 21.4 \% & N/A    & 48.8 \% & 54.1 \% & 37.0 \% & 44.8 \% & 54.1 \% \\
KVD           & 40.6 \% & 70.4 \% & 73.8 \% & N/A & 69.4 \% & 56.8 \% & \textbf{63.8} \% & 59.1\% \\
Avg. FID      & 35.5 \% & 71.2 \% & 52.0 \% & 43.5 \% & 62.4 \% & \textbf{62.7} \% & 57.6 \% & 51.2 \% \\
\hline \hline
Among raters & 79.3 \% & 77.8 \% & 84.4 \% & 74.3 \% & 83.3 \% & 69.9 \% & 72.5 \% & 74.1 \% \\\hline    
\end{tabular}
\caption{Agreement of metrics with human judgment when considering models with a fixed value for a given metric (eq.), or with spread values over a wide range (spr.). FVD is superior at judging generated videos based on subjective quality.}
\label{tbl:humaneval2}
\end{center}
\end{table*}

One important criterion for the performance of generative models is the visual fidelity of the samples as judged by human observers~\cite{bib:theis2016}.
Hence, a metric for generative models must ultimately correlate well with human judgment.
To this extent we trained several conditional video generation models, and asked human raters to compare the quality of the generated videos in different scenarios.

We trained CDNA;~\cite{bib:Finn2016:cdna}, SV2P~\cite{bib:babaeizadeh2017:sv2p}, SVP-FP~\cite{bib:denton2018:svg} and SAVP~\cite{bib:lee2018:savp} on the BAIR data set. Using the same wide range of hyper-parameters as in Section~\ref{sec:benchmark}  and by including model parameters at various stages of training we obtain over 3000 different models. 
Generated videos are obtained by combining 2 frames of context with the proceeding 14 output frames.
Following prior work~\cite{bib:babaeizadeh2017:sv2p, bib:denton2018:svg, bib:lee2018:savp} we obtain the PSNR and SSIM scores by generating 100 videos for each input context (conditioning frames) and returning the best frame-averaged value among these videos.
We consider 256 video sequences (unseen by the model) to estimate the target distribution when computing FVD.

We conduct several human studies based on different subsets of the trained models.
In particular, we select models according to two different scenarios:
\vspace{-0.1cm}
\paragraph{One Metric Equal} We try to answer the question ``if one metric is unable to distinguish between models, are those models truly equal in performance?''. We select models that are indistinguishable according to a single metric, and test if human raters and other competing metrics are able to distinguish them clearly in terms of the quality of their generated videos.
We choose 10 models having roughly equal values for a given metric that are close to the best quartile of the overall distribution of that metric.
The picked models where identical up to the first 4-5 significant digits according to the metric in question.
\vspace{-0.1cm}
\paragraph{One Metric Spread} We try to answer the question ``if according to one metric there is a clear ranking among models, do humans/other metrics agree?''. Ie., we consider to what degree models with very different scores, coincide with the subjective quality of their generated videos as judged by humans.
We choose 10 models which were equidistant between the 10\,\% and 90\,\% percentile of the overall distribution of that metric.
In this case there should be clear differences in terms of the quality of the generated videos among the models under consideration, provided that the metric is accurate.
\vspace{-0.1cm}
\paragraph{Set-up} For the human evaluation, we used 3 generated videos from each selected model.
Human raters would be shown two videos from different models, and then asked to identify which of the two looked better, or alternatively report that their quality was indistinguishable.
Each pair of compared videos was shown to up to 3 independent raters, where the third rater was only asked if the first two raters disagreed.
The raters were given no prior indication about which video was thought to be better.
We calculated the correspondence between these human ratings and the ratings determined by the various metrics under consideration.

\paragraph{Results}
The results of the human evaluation studies can be seen in Table~\ref{tbl:humaneval2}. 
In general we find that FVD is the superior choice compared to all other metrics tested.
The results obtained for \emph{eq. FVD} and \emph{spr. FVD} are of key importance as they determine how users will experience FVD in practice. 
From the \emph{spr. FVD} column we can conclude that no other metric can improve upon the ranking induced by FVD, and the \emph{eq. FVD} column tells us that no other metric can reliably distinguish between good models that are equal in terms of FVD. 
On the other hand, FVD is able to distinguish models when other metrics can not (\emph{eq. SSIM}, \emph{eq. PSNR}), agreeing well with human judgment (74.9\,\%, and 81.0\,\% agreement respectively).
Likewise FVD consistently improves on the ranking induced by other metrics (\emph{spr. SSIM}, \emph{spr. PSNR}), even though these scenarios are clearly favorable for the metric under consideration.

We find that Avg. FID performs markedly worse compared to FVD in most scenarios, except on \emph{spr. SSIM}, where it achieves slightly better performance. 
It suggests that it is preferential to judge the wide range of videos (sampled from each decile as determined by SSIM) based primarily on frame-level quality. On the other hand, when considering videos of similar quality in \emph{eq. SSIM}, we find that judging based on temporal coherence (in addition to frame-level quality) is beneficial and Avg. FID performs worse. 
KVD performs similar to FVD, although in most scenarios it performs slightly worse than FVD in terms of agreement with human judgment.
Table~\ref{tbl:humaneval2} also reports the agreement among raters.
These are computed as the fraction of the comparisons in which the first two raters agreed for a given video pair, averaged across all comparisons to obtain the final percentage. It can be seen that in most cases the raters are confident in comparing generated videos.

\subsubsection{Resolution of FVD}
While Section~\ref{sec:fvdsamplesize} demonstrates that FVD results are highly reproducible for a fixed sample size, it does not consider to what degree small differences in FVD can be considered meaningful.
Thus, human raters were asked to compare videos generated by a randomly chosen model having an FVD of 200 / 400 (base200 / base400) and generated videos by models that were 10, 20, 50, 100, 200, 300, 400, and 500 FVD points worse.
In each case we selected 5 models from the models available at these FVD scores and generated 3 videos for each model, resulting in a total of 1\,800 comparisons. For a given video comparison, raters were asked to decide which of the two videos looked better, or if they were of similar quality.
For each of these pairs, we asked up to 3 human raters for their opinion.

\begin{figure}[h!]
\begin{center}
\includegraphics[width=1.0\linewidth]{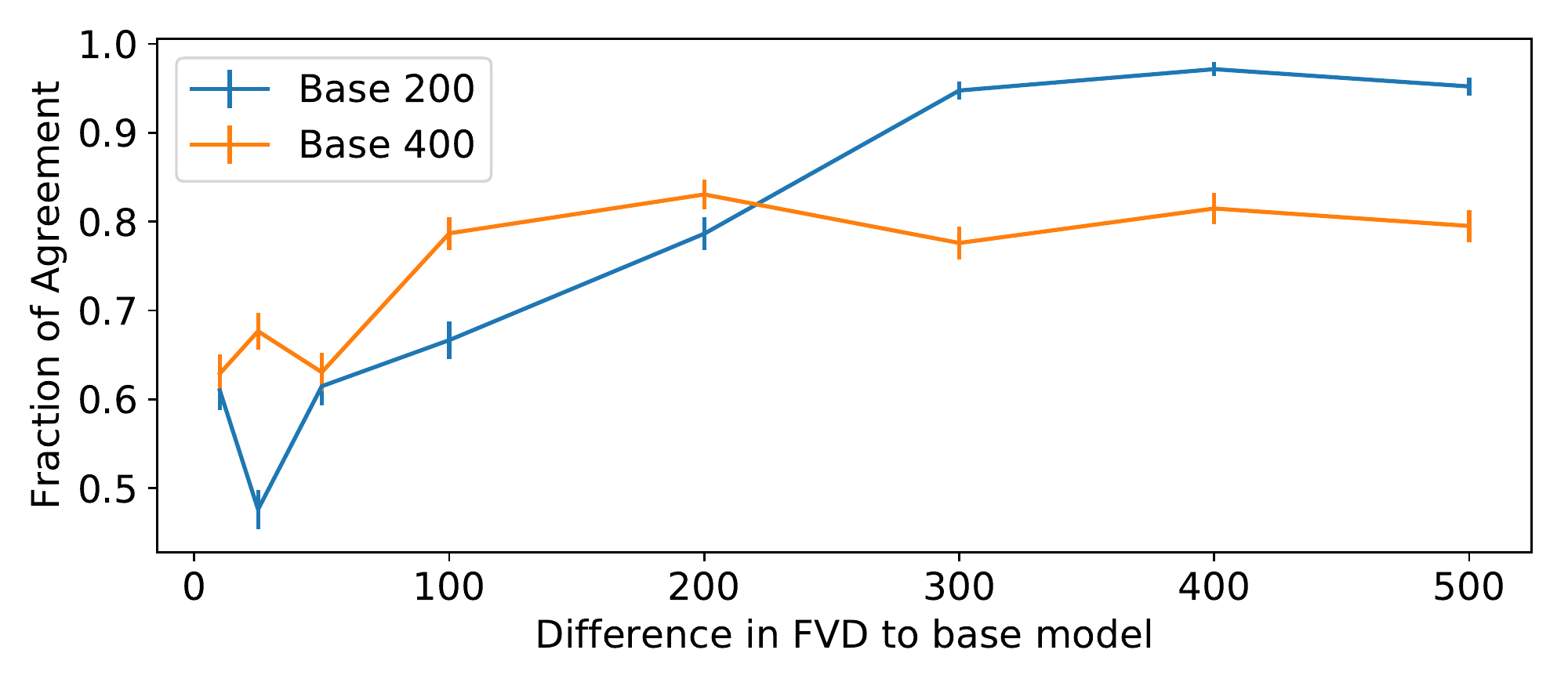}
\end{center}
\caption{Fraction of human raters that agree with FVD on which of two models is better, as a function of the difference in FVD between the models. Error bars are standard errors, and raters deciding that video pairs are of similar quality are counted as not agreeing with FVD.}
\label{fig:humaneval_resolution}
\end{figure}

In Figure~\ref{fig:humaneval_resolution} it can be seen that when the difference in FVD is smaller than 50, the agreement with human raters is close to random (but never worse), and increases rapidly once two models are more than 50 FVD points apart.
Hence, differences of 50 FVD or more typically seem to correspond to perceivable differences in the quality.

\subsection{Baseline Results on SC2 Benchmark Data Sets}\label{sec:benchmark}
\begin{table*}[t!]
\centering
\begin{tabular}{ l | r r | r r | r r | r r | r r}
\hline
\textbf{Model} & \textbf{BAIR} & \textbf{KTH} & \multicolumn{2}{c|}{\textbf{SCV-MUtB}} & \multicolumn{2}{c|}{\textbf{SCV-CMS}} & \multicolumn{2}{c}{\textbf{SCV-Brawl}} & \multicolumn{2}{c}{\textbf{SCV-RTwM}} \\
\hline
CDNA	&	296.5	&	150.8	&	486.1	&	\textbf{51.4}	&	440.8	&	515.3	&	877.1	&	1016.6 & 1089.3	&	1295.4	\\
SV2P	&	262.5	&	136.8	&	423.9	&	710.5	&	430.4	&	316.0	&	859.9	&	995.9 & 1068.7	&	\textbf{1026.1}	\\
SVP-FP	&	315.5	&	208.4	&	\textbf{276.7}	&	121.3	&	379.8	&	442.1	&	714.5	&	1240.7 & 1022.9	&	2031.4	\\
SAVP	&	\textbf{116.4}	&	\textbf{78.0}	&	479.7	&	204.4	&	\textbf{188.8}	&	\textbf{192.5}	&	\textbf{192.9}	&	\textbf{150.3} & \textbf{698.6}	&	1055.4	\\
\hline  
\end{tabular}
\caption{FVD scores on various data sets. For SCV data sets, left column is $(64 \times 64)$, right is $(128 \times 128)$.}
\label{tbl:benchmarkresults}
\end{table*}

In order to provide baseline results on our data sets, as well as to give an indication of the range of sensible values for FVD, we provide results for CDNA, SV2P, SVP-FP and SAVP on BAIR and KTH as well as all the scenarios of our SC2 benchmark suite.
Models on BAIR, MUtB, CMS and Brawl were provided 2 context frames and trained to generate 14 output frames.
For KTH, we follow the literature standard and provide 10 frames of context and train models to output 10 frames.
On RTwM we provide 2 context frames and then output up to 32 additional frames, which is the maximal length of an RTwM scenario.
For evaluation, we used 256 validation samples when calculating FVD for BAIR, 1024 samples for KTH and SCV. For SCV, each sample starts at frame 0 of the respective video, while for KTH and BAIR we extract shorter 16 or 20 frame-long consecutive subsequences from the longer videos. Hyperparameter settings can be found in the Appendix.

Table~\ref{tbl:benchmarkresults} shows the results of this benchmark. Although there are substantial differences in the quality of the generated videos (as reflected by the FVD scores), all models have similar failure modes.
This is not surprising, considering that many of these models are build around the same underlying ideas of combining CNNs, LSTMs and VAEs.
We found that the best performing models on MUtB were able to accurately synthesize videos at low resolutions, whereas weaker models sometimes failed to generate the moving unit in a higher resolution.
On the other hand, the CMS scenario caused most of the models to fail in producing accurate results.
Models were able to learn that shards would disappear from the map, yet they failed both at modeling the moving units themselves, and at capturing the correct sequence with which the mineral shards were meant to be disappearing.
On the Brawl map, the models were unable to capture all of the game units participating in the brawl, and instead resorted to generate larger blurry blobs.
Perhaps surprisingly, neither model was able to succeed at modeling the RTwM scenario.
Whether this is due to the longer video sequences that are required to be considered or inherent to the complexity of the scenario remains a question for further research.
Please see the Appendix for examples of generated samples ($128 \times 128$) that are representative of various behaviors.

\subsection{Correlation of FVD with SSIM and PSNR}
As a final experiment we evaluate the degree to which FVD correlates with SSIM and PSNR.
Using all the models from Section~\ref{sec:benchmark}, we end up with more than 20\,000 different models.
These can be used to obtain an extensive range of potentially different generated videos that one could encounter in training generative models of video.
Overall, we find that the correlation between SSIM and PSNR is generally very high (Pearson's r= 0.730, Kendall's $\tau$=0.648), which is unsurprising given that both metrics are based on the same frame-by-frame measurements. 
There was a weaker, yet still statistically significant correlation between SSIM and FVD (r=-0.640, $\tau$=-0.189), which indicates that SSIM does to some extend pick up to the same types of defects that FVD also detects.
The correlation between FVD and PSNR was rather weak (r=-0.278, $\tau$=-0.007).

\section{Conclusion}
We introduced the Fr\'{e}chet Video Distance (FVD), a new evaluation metric for generative models of video, and an important step towards better evaluation of models for video generation.
By design, the currently favored metrics, SSIM and PSNR, can only account for the quality of individual frames, and are restricted to the case in which ground-truth sequences are available.
In contrast, FVD can even be used in situations where this is not the case, such as unconditional video generation via Generative Adversarial Networks.
Our experiments confirm that FVD is accurate in evaluating videos that were modified to include static noise, and temporal noise.
More importantly, a large scale human study among generated videos from several recent generative models reveals that FVD consistently outperforms SSIM and PSNR in agreeing with human judgment.

We also presented the StarCraft 2 Videos (SCV) benchmark, which introduces several new data sets of different complexity that test long-term memory and relational reasoning.
In our evaluation of state of the art models for video generation, we find that these challenges are indeed still open.
We hope that by isolating (independent) challenges in a visually simpler setting, it becomes more easy for researchers to propose, test, and analyze potential solutions to these specific problems.
We believe that FVD and SCV will greatly benefit research in generative models of video in providing a well tailored, objective measure of progress.

{\small
\bibliographystyle{ieee}
\bibliography{references}
}

\onecolumn
\appendix
\clearpage

\section{Noise Study}\label{sec:noisestudy-details}

We conduct the noise study on HMDB~\cite{bib:kuehne2013:hmdb51}, BAIR~\cite{bib:ebert2017:bair}, and Kinetics-400~\cite{bib:kay2017:kinetics}.
A total of $90\%$ of the available samples (train and test) were used to perform the comparison. 
A mapping of the different noise intensities to the parameter values of the various noise types that we consider can be seen in Table~\ref{tbl:noisemap}.

\begin{table*}[hb!]
\begin{center}
\begin{tabular}{l|c|c|c|c|c|c|c} \hline
\textbf{Noise type} & \textbf{Parameter} & \textbf{Int. 1} & \textbf{Int. 2} & \textbf{Int. 3} & \textbf{Int. 4} & \textbf{Int. 5} & \textbf{Int. 6} \\ \hline
Black rectangle & size relative to image & 15\% & 30\% & 45 \% & 60 \% & 75 \% & N/A \\
Gaussian blur           & sigma of Gaussian kernel & 1       & 2 & 3 & 4 & 5 & N/A \\
Gaussian noise           & percentage of noise in convex combination & 15\% & 30\% & 45 \% & 60 \% & 75 \% & N/A \\
Salt \& Pepper           & probability of applying `salt' or `pepper' & 0.1 & 0.2 & 0.3 & 0.4 & 0.5 & N/A \\
\hline \hline
Local swap & number of swaps & 4 & 8 & 12 & 16 & 20 & 24 \\
Global swap           & number of swaps & 4 & 8 & 12 & 16 & 20 & 24 \\
Interleaving           & number of sequences & 2 & 3 & 4 & 5 & 6 & N/A \\
Switching           & number of frames until switch & 1 & 2 & 3 & 4 & 5 & N/A \\ \hline
\end{tabular}
\caption{An overview of the different noise intensities used for different noise types.}
\label{tbl:noisemap}
\end{center}
\end{table*}

\begin{figure*}[hb!]
\begin{center}
\includegraphics[width=1.0\linewidth]{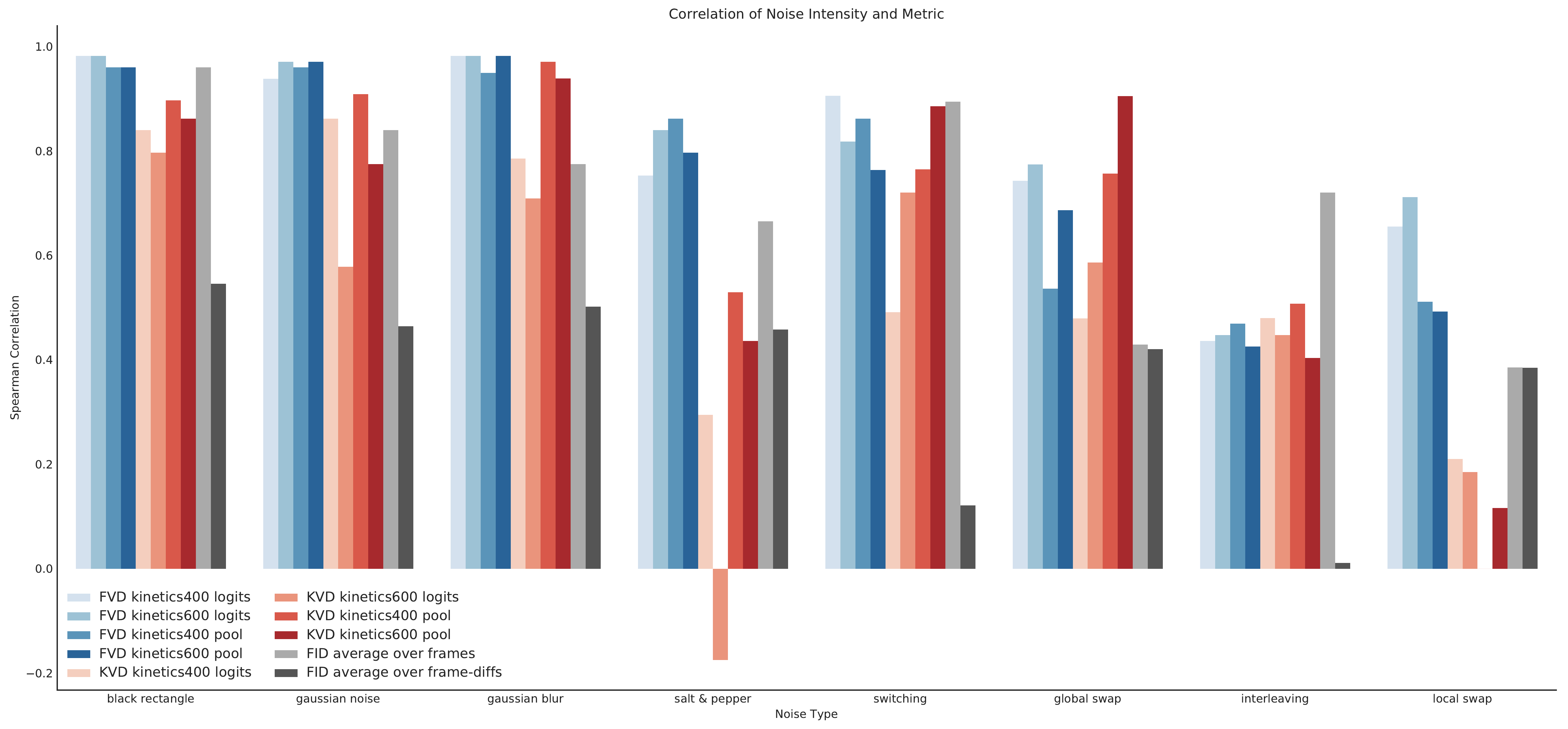}
\end{center}
\caption{Correlation of the noise intensity and the metric measurements.}
\label{fig:noiseintensity}
\end{figure*}

Figure~\ref{fig:noiseintensity} provides an overview of the correlation of various implementations of FVD (and an FID-based baseline) with the sequence of noise intensities.
It can be seen that the logits of the I3D model trained on the Kinetics 400 dataset correlates well with the noise intensities, across a variety of noise types.

\newpage
\section{SCV Data Generation}\label{sec:scv-datageneration}
An overview of the stochastic elements and main objectives of each scenario is listed in Table~\ref{tbl:scv_scenarios}.

Each data set in SCV consists of 10\,000 training, 2\,000 validation and 2\,000 test set videos of game play of an agent playing a custom StarCraft 2 scenario.
Scenarios were created using the Starcraft 2 Editor, and agents were implemented in the SC2LE framework.
All agents are deterministic, and randomness is controlled by the environment, as implemented by the scenario. 
Actions are provided in the form of visual cues (right-mouse clicks), which guide a model in predicting the next frame e.g. in MUtB, the first frame indicates the direction the unit will move to, or in CMS, it shows which crystal a unit will target next.

Data is obtained in two phases. 
First a set of replay files is created by having an agent ``play'' a scenario in SC2LE.
These encode a sequence of deterministic states in the environment, allowing the same content to be rendered at different resolutions. 
A sequence of states (agent interacting with the environment) is terminated once a particular `termination condition' has been fulfilled. 
Different scenarios have different termination conditions:
\begin{itemize}
    \item MUtB: scenario ends when a unit reaches the border
    \item CMS: scenario lasts for 2 in-game minutes
    \item Brawl: scenario ends when one of the armies is victorious or 2 in-game minutes pass
    \item RTwM: scenario ends when all units are unloaded at the final beacon
\end{itemize}

Since these scenarios have widely different lengths, we rendered them at different speeds: in MUtB, we recorded every $6^{th}$ frame, in CMS and Brawl every $4^{th}$ frame, and in RTwM every $8^{th}$ frame.
We additionally always skipped the first 2 frames of each video, as these first frames are being rendered before the scenario is fully initialized.
The stochasticity in each scenario results in replays (and corresponding videos) having different lengths. 
The generated data sets contain between 11-27 frames on MUtB, up to 99 frames on Brawl, and exactly  99/32 frames on CMS/RTwM respectively. We subsample shorter videos of the required length when training.

\begin{table*}[hb!]
\centering
\begin{tabular}{ l | l | l }
\hline
\textbf{Scenario} & \textbf{Stochastic Elements} & \textbf{Main Challenge} \\
\hline
MUtB & unit type, unit color, moving direction      & `Unit test', reproducing movement animations \\
CMS & placement of crystals, path through crystals  & Learning complex action sequences and modelling movement paths \\
Brawl & army side, unit types, unit positions & Modelling many moving entities, and their interactions \\
RTwM & number of units, unit types, beacon location  & Remembering events over many time steps \\
\hline  
\end{tabular}
\caption{An overview of the properties of the SCV scenarios.}
\label{tbl:scv_scenarios}
\end{table*}

\subsection{Scenario Parameters}
\label{sec:scenario-params}
Each SCV scenario is implemented in the form of an ``SC2Map'' file that can be read by the official StarCraft2 Editor.
The scenario specifies how to initialize the map, units, and observes the game state to test the termination condition.
In the process of designing each SCV scenario we created several hyperparameters whose values can be changed to adjust the complexity of the scenario while retaining the core task.
For example in Brawl, increasing or decreasing the number of units in each army will affect the number of interactions that take place.
Likewise, in RTwM the task can be made more difficult by adding additional beacons that have to be visited before unloading.
In the following we describe the default values of the hyperparameters that we used to create SCV

\paragraph{Move Unit to Border}
A single unit spawns in the center of the map, and a beacon is placed along the outside border of the map (not visible on screen) that the unit moves towards.
The unit is randomly chosen among 6 types (Marine, Siege Tank, Drone, Zergling, Colossus, Archon) and colored randomly according to 4 colors (red, blue, green, yellow).

\paragraph{Collect Mineral Shards}
Two units spawn at random locations in the map.
20 mineral shards are placed randomly in the map with a minimum distance from the units.
The unit types are currently fixed (always marines), and we do not apply any coloring.
Each unit is repeatedly tasked to move to the nearest mineral shard to collect it.

\paragraph{Brawl}
Two armies (one Zerg, one Terran) face off.
The army spawn locations are vertical strokes along the left and right border, which are randomly assigned to each army.
Each army includes a fixed number of units (currently 9) that are spawned in a random location within the corresponding army spawn location.
Each unit is randomly chosen among 6 types (Roach, Hydralisk, Ultralisk, Baneling, Brood Lord, Queen for Zerg and Marine, Marauder, Thor, Battlecruiser, Siege Tank, Banshee for Terran).
The Terran army attacks, repeatedly targeting the nearest unit.

\paragraph{Road Trip with Medivac}
A single transporation unit (Medivac) is spawned randomly in the center area of the map.
Up to 4 units (randomly selected between 1-4) are spawned randomly in an area surrounding the medivac.
Each unit is randomly chosen among 6 types (Marine, SCV, Marauder, Reaper, Hellion, Ghost) and are assigned mixed colors (red, green, blue, yellow). 
Two beacons are spawned randomly in the map, while avoiding the center locations.
The Medivac picks up each of the units in a greedy closest-first fashion.
It then visits each of the beacons, before unloading the same units again upon visiting the final beacon.

\section{Benchmark Hyperparameters}\label{sec:benchmarkhyparparams}
We used the Tensor2Tensor~\cite{bib:Vaswani2018:t2t} implementation of each model and the default parameters unless otherwise stated.
Suitable hyper-parameters were obtained using a grid-search over the learning rate ($10^{-3}, 10^{-4}$ and $10^{-5}$), and the trade-off between reconstruction loss and KL divergence ($\beta$: $10^{-6}, 10^{-5}, 10^{-6}, 10^{-3}$) for VAE-based models.
For SV2P, we followed the $\beta$ annealing schedule as outlined in~\cite{bib:babaeizadeh2017:sv2p}.
For SAVP, we additionally tuned the GAN-loss and the GAN-VAE-loss by searching for suitable values on a logarithmic scale between $10^{-6}$ and $10^{-3}$.
All models were trained for a total of 300\,000 update steps.

\newpage
\section{Examples of Video Models on SCV}\label{sec:benchmarkexamples}
In the following examples, the top row shows a video sequence from the data set $(128 \times 128)$, and the row below shows the video generated by a model that was conditioned on the first two frames of that sequence.
Compare to the quality of the generated samples to the FVD scores on the right columns in \autoref{tbl:benchmarkresults}.

\begin{figure*}[h!]
\begin{center}
\includegraphics[width=1.0\linewidth]{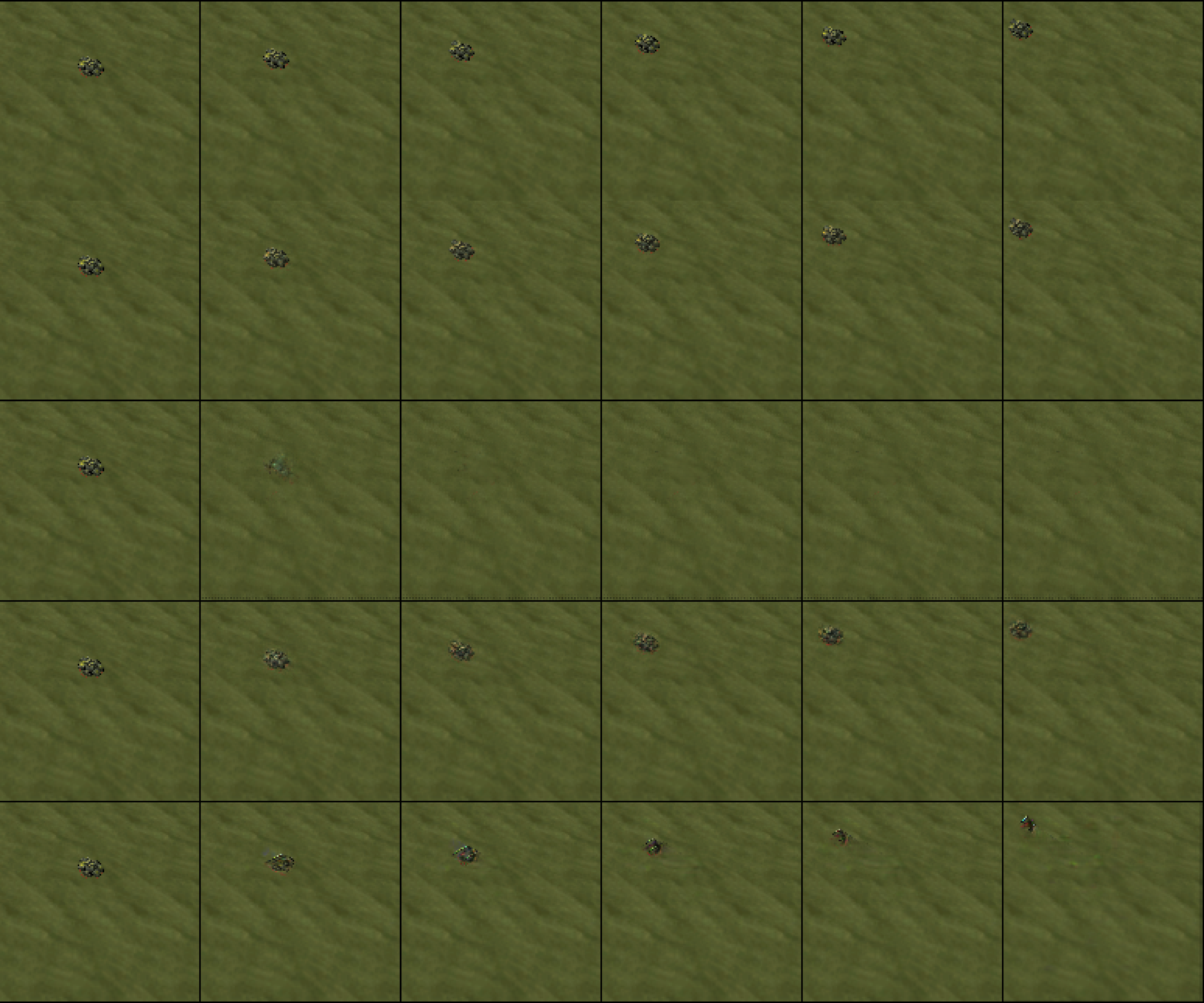}
\end{center}
\caption{Examples from Move Unit to Border in $(128 \times 128)$ resolution. Top to Bottom: Original video, CDNA, SV2P, SVP-FP, SAVP}
\end{figure*}

\begin{figure*}[h!]
\begin{center}
\includegraphics[width=1.0\linewidth]{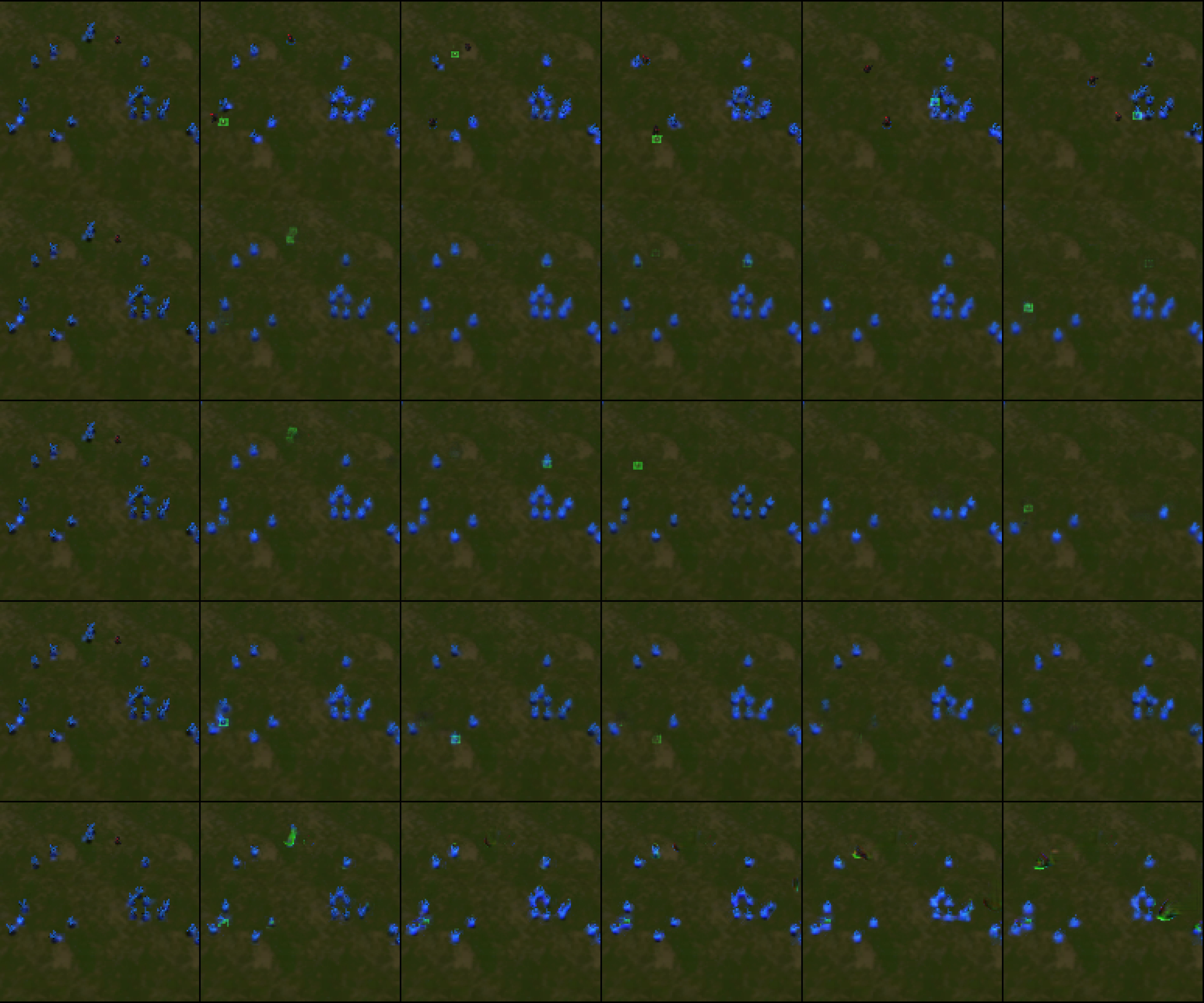}
\end{center}
\caption{Examples from Collect Mineral Shards in $(128 \times 128)$ resolution. Top to Bottom: Original video, CDNA, SV2P, SVP-FP, SAVP}
\end{figure*}

\begin{figure*}[h!]
\begin{center}
\includegraphics[width=1.0\linewidth]{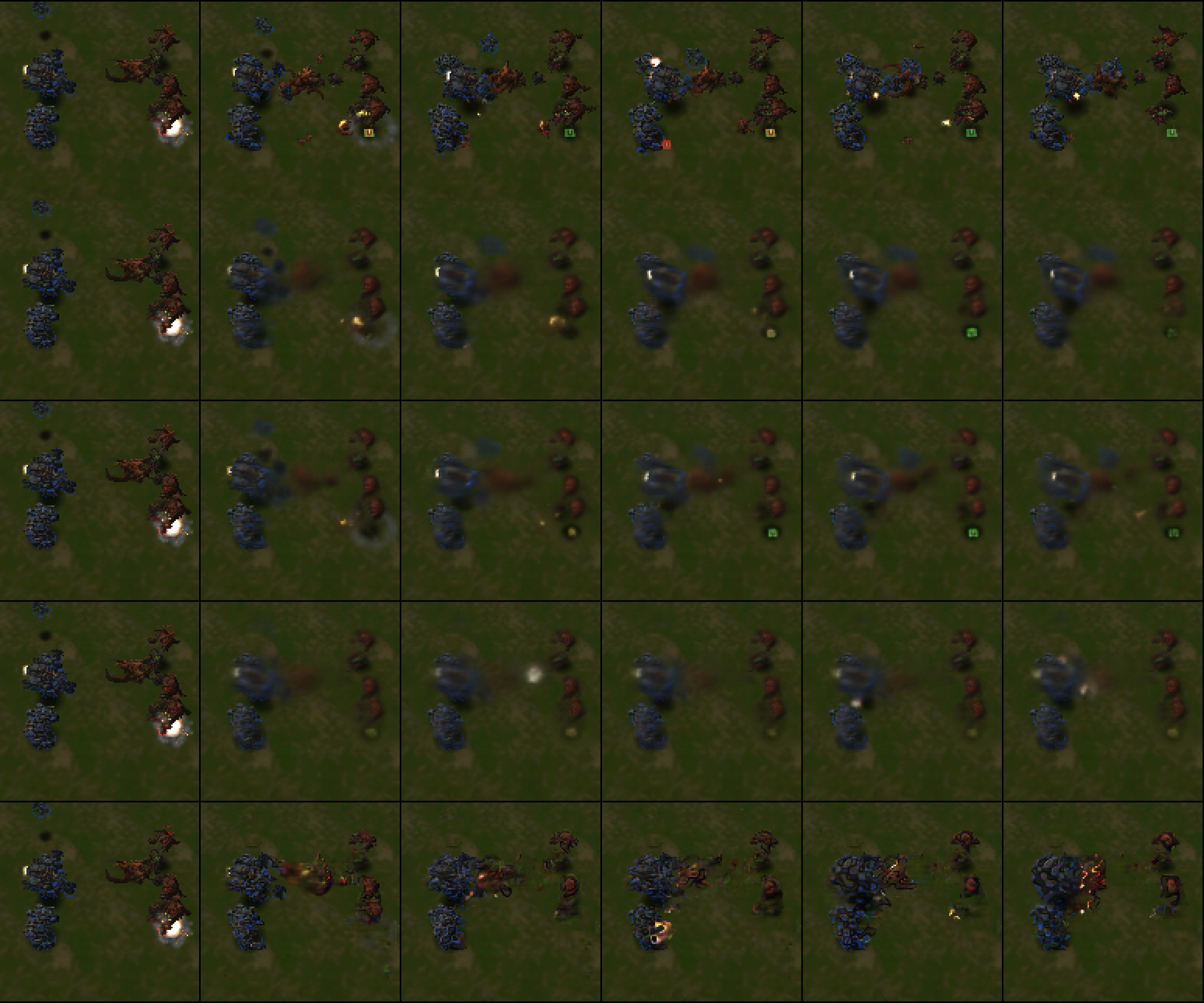}
\end{center}
\caption{Examples from Brawl. Top to Bottom in $(128 \times 128)$ resolution: Original video, CDNA, SV2P, SVP-FP, SAVP}
\end{figure*}

\begin{figure*}[h!]
\begin{center}
\includegraphics[width=1.0\linewidth]{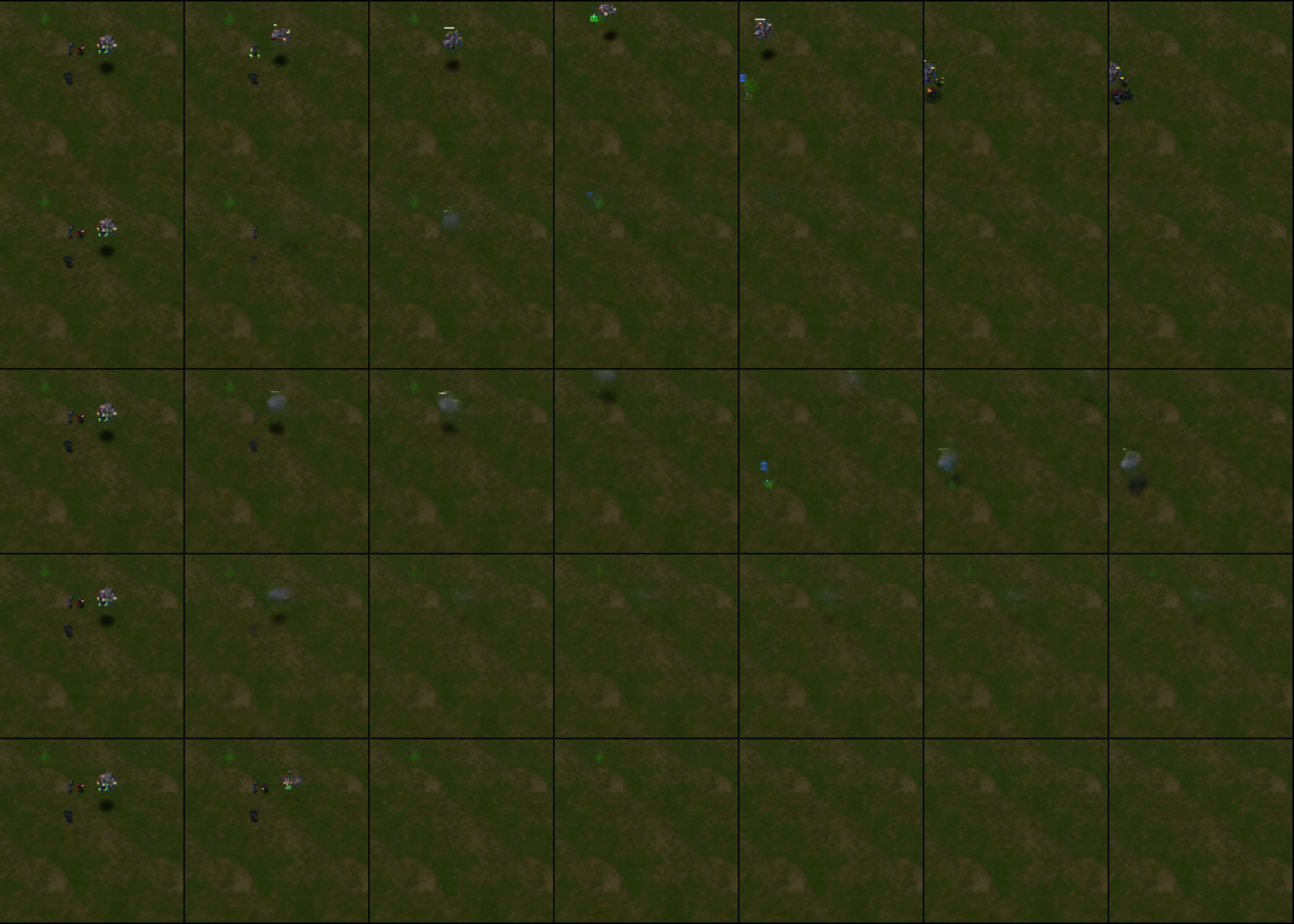}
\end{center}
\caption{Examples from Road Trip with Medivac in $(128 \times 128)$ resolution. Top to Bottom: Original video, CDNA, SV2P, SVP-FP, SAVP}
\end{figure*}

\end{document}